%% file: aaai2021.tex
\def\year{2021}\relax
\documentclass[letterpaper]{article} 
\usepackage{aaai21}  
\usepackage{times}  
\usepackage{helvet} 
\usepackage{courier}  
\usepackage[hyphens]{url}  
\usepackage{graphicx} 
\usepackage{graphicx}  
\usepackage{natbib}  
\usepackage{caption} 
\usepackage{amsmath}
\usepackage{amsfonts}
\usepackage{booktabs}
\usepackage{physics}
\usepackage{amssymb}
\title{Context-Guided BERT for Targeted Aspect-Based Sentiment Analysis}
\author{

    Zhengxuan Wu \textsuperscript{\rm 1},
    Desmond C. Ong \textsuperscript{\rm 2, 3}
    \\

}
\affiliations{
    \textsuperscript{\rm 1} Symbolic Systems Program, Stanford University \\
    \textsuperscript{\rm 2} Department of Information Systems and Analytics, National University of Singapore \\
    \textsuperscript{\rm 3} Institute of High Performance Computing, Agency for Science, Technology, and Research, Singapore \\
    wuzhengx@stanford.edu, dco@comp.nus.edu.sg

}

\usepackage{array}
\usepackage{multirow}
\usepackage{color}

\begin{document}

\maketitle

\begin{abstract}
Aspect-based sentiment analysis (ABSA) and Targeted ABSA (TABSA) allow finer-grained inferences about sentiment to be drawn from the same text, depending on context. For example, a given text can have different targets (e.g., neighborhoods) and different aspects (e.g., price or safety), with different sentiment associated with each target-aspect pair.
In this paper, we investigate whether adding context to self-attention models improves performance on (T)ABSA. We propose two variants of Context-Guided BERT (CG-BERT) that learn to distribute attention under different contexts.  
We first adapt a context-aware Transformer to produce a CG-BERT that uses context-guided \emph{softmax}-attention. Next, we propose an improved Quasi-Attention CG-BERT model that learns a compositional attention that supports subtractive attention. We train both models with pretrained BERT on two (T)ABSA datasets: SentiHood and SemEval-2014 (Task 4). Both models achieve new state-of-the-art results with our QACG-BERT model having the best performance. Furthermore, we provide analyses of the impact of context in the our proposed models. Our work provides more evidence for the utility of adding context-dependencies to pretrained self-attention-based language models for context-based natural language tasks.
\end{abstract}

\input{1_introduction}

\input{2_relatedwork}

\input{3_approach}

\input{4_experiment}

\input{5_conclusion}

\input{6_acknowledgement}

\begin{quote}
\begin{small}
\bibliography{aaai2021}
\end{small}
\end{quote}

\end{document}

%% file: 1_introduction.tex
\section{Introduction}
People are living more of their lives online, both on social media and on e-commerce platforms, and this trend was exacerbated by the recent need for social distancing during the Covid-19 pandemic. Because people are using online review platforms like Yelp and delivery platforms more frequently, understanding the types of emotional content generated on such platforms could yield business insights or provide personalized recommendations~\cite{kang2012senti}. To this end, Sentiment Analysis techniques have been applied to understand the emotional content generated on microblogs~\cite{kouloumpis2011twitter, severyn2015twitter}, online reviews (e.g., movie and restaurant reviews)~\cite{socher2013recursive, kiritchenko2014nrc}, narratives~\cite{wu2019attending, ong2019modeling} and other online social media~\cite{lwin2020global}. 

However, user-generated reviews contain more complex information than just a single overall sentiment. A review of an upscale neighborhood (for potential renters or home-buyers) may praise the safety but express incredulity at the price. Identifying the different \emph{aspects} (e.g., price, safety) embedded within a given text, and their associated sentiment, has been formalized as a task called Aspect-Based Sentiment Analysis (ABSA)~\cite{pontiki2016semeval, poria2016deeper}. Targeted ABSA (TABSA) is a more general version of ABSA, when there are multiple targets in a review, each with their associated aspects. For example, given a review of neighborhoods: ``\emph{LOC1} area is more expensive but has a better selection of amenities than in \emph{LOC2}" (where \emph{LOC1} and \emph{LOC2} are specially masked tokens), we note that the sentiment depends on the specific target (\emph{LOC1} or \emph{LOC2}) and their aspect. The sentiment towards the \textbf{price} of \emph{LOC1} may be negative---and may be more important to a price-conscious student---but positive in terms of \textbf{convenience}, while the sentiment towards \emph{LOC2}'s aspects are reversed.

Research using neural models for (T)ABSA has mainly focused on using deep neural networks such as RNNs or attention-gated networks to generate context-dependent sentence embeddings~\cite{saeidi2016sentihood, tang2016aspect, wang2016attention, chen2017recurrent, ma2018targeted, liu2018recurrent}. 
Recently, with the advent of powerful self-attention models like the Transformer and BERT, \citet{sun2019utilizing} and \citet{li2019exploiting} both applied pretrained BERT models to (T)ABSA, and showed promising performance improvements. However, these approaches simply used a pretrained BERT model as a blackbox: either via using BERT as an embedding layer or appending the aspect to the input sentence. 

We propose to improve the BERT model architecture to be \emph{context-aware}. A context-aware model~\cite{yang2019context} should distribute its attention weights appropriately under different contexts---in (T)ABSA, this means specific targets and aspects. Additionally, by incorporating context into the calculation of attention weights, we aim to enrich the learnt hidden representations of the models. 
Specifically, we propose two methods to integrate context into the BERT architecture: (1) a Context-Guided BERT (CG-BERT) model adapted from a recent context-aware self-attention network~\cite{yang2019context}, which we apply to (T)ABSA; and (2) a novel Quasi-Attention  Context-Guided BERT (QACG-BERT) model that learns quasi-attention weights---that could be negative---in a compositional manner and which enables subtractive attention, that is lacking in softmax-attention~\cite{tay2019compositional}. In particular, our contribution is three-fold~\footnote{\url{https://github.com/frankaging/Quasi-Attention-ABSA}}:
\begin{enumerate}
  \item We extend a recently-proposed context-aware self-attention network \cite{yang2019context} to the (T)ABSA task by formulating a Context-Guided BERT model (CG-BERT).
  \item We propose a new Quasi-Attention Context-Guided BERT model (QACG-BERT) that achieves new state-of-the-art (SOTA) results on two (T)ABSA datasets.
  \item We analyze how context influences the self-attention and decisions of our models.
\end{enumerate}

%% file: 2_relatedwork.tex
\section{Background and Related Work}

\begin{figure}
    \centering
    \includegraphics[width=0.48\textwidth]{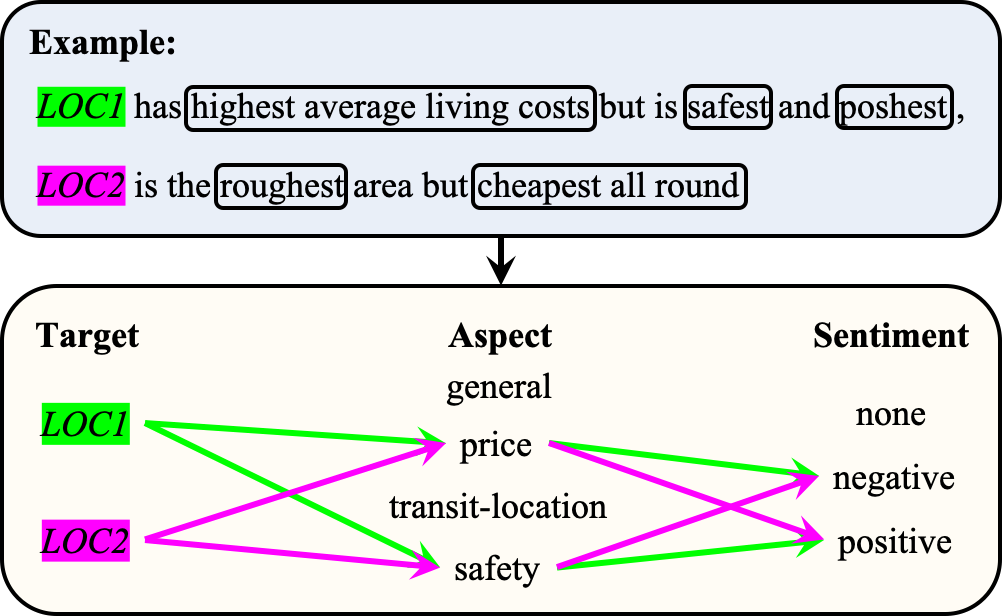}
    \caption{A labelled example from the SentiHood dataset, which compares the \textbf{price} and \textbf{safety} of two locations. 
    The aspects not mentioned in the text are given the label \emph{none}.
    }
    \label{tab:example}
\end{figure}

\subsection{Self-attention Networks}

Self-attention networks, exemplified in the Transformer~\cite{vaswani2017attention}, have become the \emph{de facto} go-to neural models for a variety of NLP tasks including machine translation~\cite{vaswani2017attention}, language modeling~\cite{liu2018learning, dai2019transformer} and sentiment analysis~\cite{shen2018disan, wu2019attending}. BERT, a popular and successful variant of the Transformer, has successfully been applied to various NLP tasks~\cite{reddy2019coqa, devlin2019bert}. In addition, previous studies have also shown some evidence that self-attention weights may learn syntactic~\cite{hewitt2019structural} and semantic~\cite{wu2020structured} information.


\subsection{Formulation of \emph{Quasi}-attention}
Attention has been successfully applied to many NLP tasks. Various forms of attention are proposed including additive attention~\cite{bahdanau2015neural}, dot-product attention~\cite{luong2015effective} and scaled dot-product attention used in self-attention~\cite{vaswani2017attention}. Many of them rely on a \emph{softmax} activation function to calculate attention weights for each position. As a result, the output vector is in the convex hull formed by all other hidden input vectors, preventing the attention gate from learning subtractive relations. \citet{tay2019compositional} recently proposed a way to overcome this limitation by allowing attention weights to be negative (``quasi" attention), which allows input vectors to add to ($+$1), not contribute to ($0$), and even subtract from ($-$1) the output vector.

\subsection{Aspect-based Sentiment Analysis}
Early research in ABSA introduced classic benchmark datasets and have proposed many baseline methods including lexicon-based and pre-neural classifiers~\cite{pontiki2014semeval, kiritchenko2014nrc, pontiki2015semeval, pontiki2016semeval}. Since the debut of recurrent neural networks, various RNNs have been developed to generate aspect-aware sentence embeddings and sentiment labels~\cite{tang-etal-2016-effective, chen2017recurrent, li2018transformation}. Likewise, researchers have also adapted CNNs~\cite{xue2018aspect, huang2018parameterized}, recursive neural networks~\cite{nguyen2015phrasernn}, aspect-aware end-to-end memory networks~\cite{tang2016aspect} and cognitively inspired deep neural networks~\cite{lei2019human} to generate aspect-aware sentence embeddings. 

Motivated by attention mechanisms in deep learning models, many recent ABSA papers have integrated attention into neural models such as RNNs~\cite{wang2016attention, chen2017recurrent, liu2017attention, he2018exploiting}, CNNs~\cite{zhang2019aspect}, and memory networks~\cite{ma2017interactive, majumder2018iarm, liu2018recurrent} to learn different attention distributions for aspects and generate attention-based sentence embeddings. Most recently, self-attention-based models such as BERT have been applied to ABSA, by using BERT as the embedding layer~\cite{song2019attentional, yu2019adapting, lin2019deep}, or fine-tuning BERT-based models with an ABSA classification output layer~\cite{xu2019bert}. These papers show that using BERT brings significant performance gains in ABSA.

\subsection{Targeted Aspect-based Sentiment Analysis}
Building on ABSA, \citet{saeidi2016sentihood} proposed a generalized TABSA task (with multiple potential targets) with a new benchmark dataset and LSTM-based baseline models. Various neural models have been proposed for TABSA such as a memory network with delayed context-aware updates~\cite{liu2018recurrent} and interaction-based embedding layers to generate context-aware embeddings~\cite{liang2019context}. Researchers have also tried to integrate attention mechanism with LSTMs to predict sentiment for target-aspect pairs~\cite{ma2018targeted}. With the recent success of BERT-based models, various papers have used BERT to generate contextualized embeddings for input sentences, which are then used to classify sentiment for target-aspect pairs~\cite{huang2019syntax, hu2019learning}. More recent papers have fine-tuned BERT for TABSA either by (i) constructing auxiliary sentences with different pairs of targets and aspects or (ii) modifying the top-most classification layer to also take in targets and aspects~\cite{rietzler2020adapt, sun2019utilizing, li2019exploiting}. To the best of our knowledge, no work has been published on modifying the BERT architecture for TABSA tasks. Instead of keeping BERT as a blackbox, we enable BERT to be context-aware by modifying its neural architecture to account for context in its attention distributions.

\begin{figure*}[!htb]
\minipage{0.25\textwidth}
  \includegraphics[width=\linewidth]{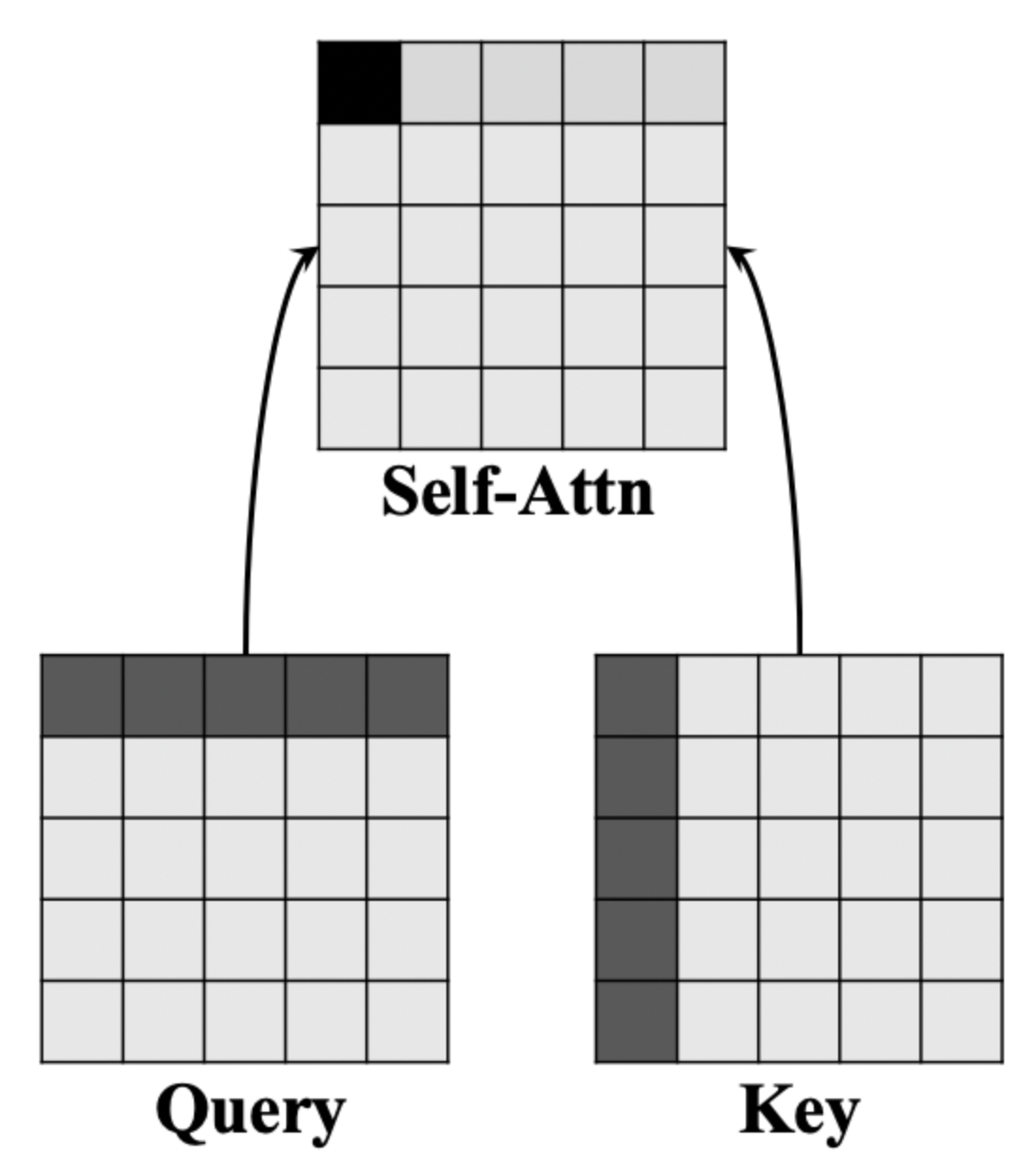}
  \caption*{(a) BERT}\label{fig:awesome_image1}
\endminipage\hfill
\minipage{0.26\textwidth}
  \includegraphics[width=\linewidth]{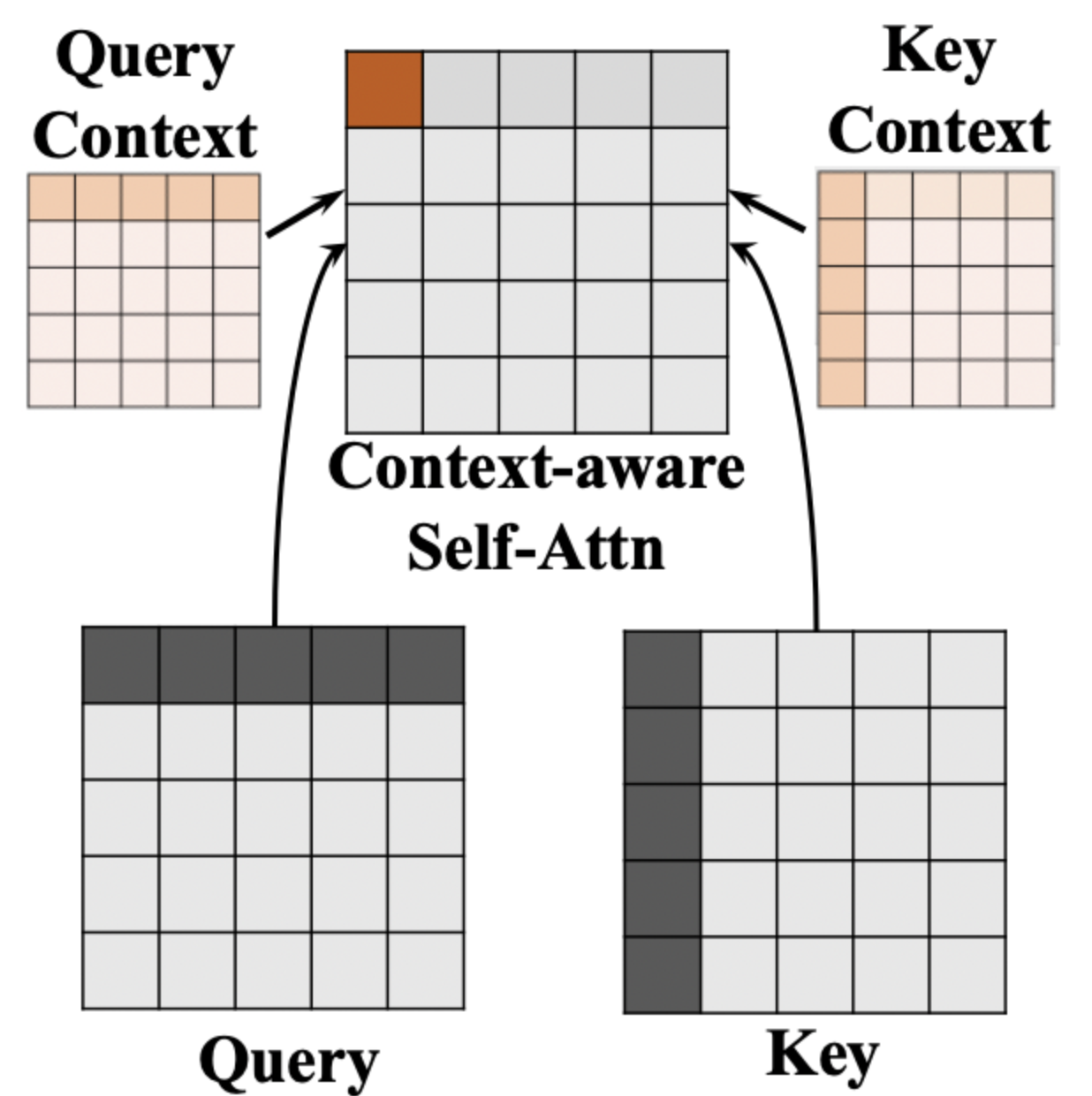}
  \caption*{(b) CG-BERT}\label{fig:awesome_image2}
\endminipage\hfill
\minipage{0.38\textwidth}%
  \includegraphics[width=\linewidth]{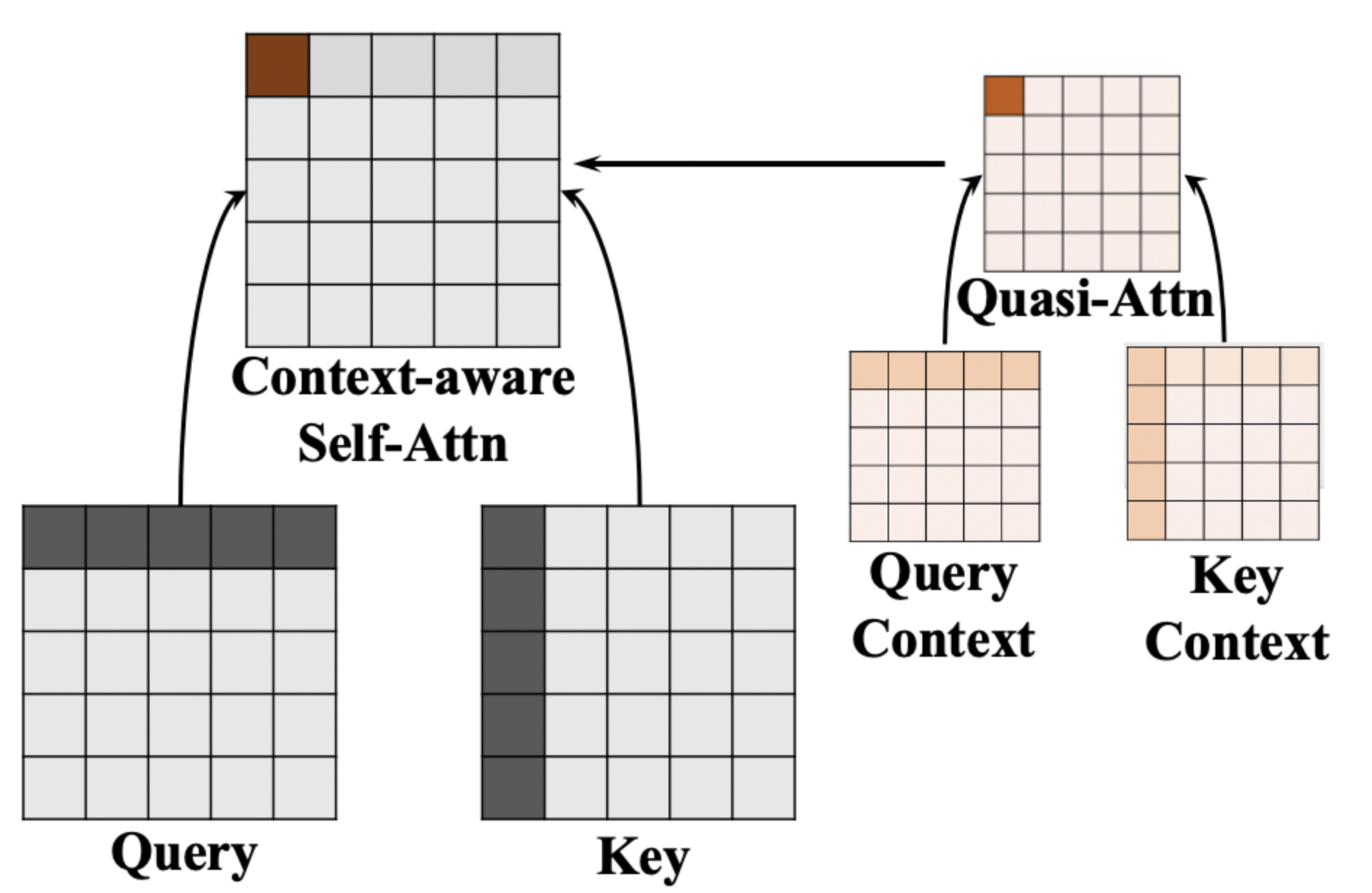}
  \caption*{(c) QACG-BERT}\label{fig:awesome_image3}
\endminipage
\caption{Illustration of the proposed models. (a) The vanilla self-attention network (e.g., BERT) calculates attention weights using the query and key matrices without considering context. (b) The CG-BERT model modifies query and key matrices using context, and then calculate attention weights as in (a). (c) The QACG-BERT model calculates attention weights by combining vanilla attention weights as in (a) with quasi-attention weights calculated using a separate pair of query and key matrices for context. Colors in the grids illustrate matrix operations.}
\label{fig:model-arch}
\end{figure*}

%% file: 3_approach.tex
\section{Approach}
We start by defining both TABSA and ABSA tasks, before we introduce our context-guided BERT models (Fig.~\ref{fig:model-arch}). First, we describe the context-guided BERT model CG-BERT that uses \emph{softmax}-attention, originally proposed by~\citet{yang2019context}, and the modifications we made to tailor it to the (T)ABSA task. Second, we propose a new neural architecture QACG-BERT that uses quasi-attention adapted from~\cite{tay2019compositional}. Lastly, we describe our methods to formulate our context matrices, and to integrate with pretrained BERT weights.

\subsection{TABSA Task}
We formulate the Sentihood dataset as a TABSA task. Given a sentence $s$ with a sequence of words $\{ w_{1}, w_{2}, ... w_{n}\}$\footnote{We append a classifier token (i.e., [CLS]) in the beginning of each input sentence as in the BERT model~\cite{devlin2019bert}.}, where some words are target pronouns $\{ w_{i}, ..., w_{j}\}$ 
from a fixed set $T$ of $k$ predefined targets $\{ t_{1}, ..., t_{k}\}$, the goal is to predict sentiment labels for each aspect associated with each unique target mentioned in the sentence. Following the setup in the original Sentihood paper~\cite{saeidi2016sentihood}, given a sentence $s$, a predefined target list $T$ and a predefined aspect list $A = $\{\textit{general}, \textit{price}, \textit{transit-location}, \textit{safety}\}, the model predicts a sentiment label $y \in $\{\textit{none}, \textit{negative}, \textit{positive}\} for a given pair of $\{(t, a):(t \in T, a \in A)\}$. Note that the model predicts a single sentiment label for each unique target-aspect pair in a sentence. We show an example of TABSA in Fig.~\ref{tab:example}.

\subsection{ABSA Task}
We use the SemEval-2014 Task 4 dataset~\cite{pontiki2014semeval} to formulate an ABSA task: Given a sentence, we predict a sentiment label $y \in $\{\textit{none}, \textit{negative}, \textit{neutral}, \textit{positive}, \textit{conflict}\} for each aspect $\{a : (a \in A)\}$ with a predefined aspect list $A = $\{\textit{price}, \textit{anecdotes}, \textit{food}, \textit{ambience}, \textit{service}\}.


\subsection{Context-Guided \emph{Softmax}-attention}
Our Context-Guided BERT (CG-BERT) model is based on the context-aware Transformer model proposed by~\citet{yang2019context}, which we adapted to the (T)ABSA task. Multi-headed self-attention~\cite{vaswani2017attention} is formulated as:
\begin{align}
    \mathbf{A}_{\text{Self-Attn}}^{h} = \text{softmax}\left(\frac{ \mathbf{Q}^{h} {\mathbf{K}^{h}}^{T}}{\sqrt{d_{h}}} \right) \label{attn-self-attn}
\end{align}
where $\mathbf{Q}^h \in \mathbb{R}^{n \times d}$ and $\mathbf{K}^h \in \mathbb{R}^{n \times d}$ are query and key matrices indexed by head $h$, and ${\sqrt{d_{h}}}$ is a scaling factor. We integrate context into BERT by modifying $\mathbf{Q}$ and $\mathbf{K}$ matrices of the original BERT model~\cite{devlin2019bert}:
\begin{gather}
 \begin{bmatrix} \mathbf{\hat{Q}}^{h} \\ \mathbf{\hat{K}}^{h} \end{bmatrix}
 =
  \left( 1 -   \begin{bmatrix}
  \lambda_{Q}^{h} \\ 
  \lambda_{K}^{h}
   \end{bmatrix} \right) 
  \begin{bmatrix}
  \mathbf{Q}^{h} \\
  \mathbf{K}^{h}
   \end{bmatrix} 
 + 
 \begin{bmatrix}
  \lambda_{Q}^{h} \\ 
  \lambda_{K}^{h}
   \end{bmatrix} 
   \left(
   \mathbf{C}^{h}   
   \begin{bmatrix}
    \mathbf{U}_{Q} \\
    \mathbf{U}_{K}
   \end{bmatrix} 
   \right) \label{ctx-self-attn}
\end{gather}
where $\mathbf{C}^{h} \in \mathbb{R}^{n \times d_{c}}$ is the context matrix for each head and defined in Sec.~\ref{sec:context}, $\{ \lambda_{Q}^{h}, \lambda_{K}^{h} \} \in \mathbb{R}^{n \times 1}$ are learnt context weights, and $\{\mathbf{U}_{Q}, \mathbf{U}_{K}\} \in \mathbb{R}^{d_{c} \times d_{h}}$ are weights of linear layers used to transform input context matrix $\mathbf{C}_{h}$. The modified $\mathbf{\hat{Q}}$ and $\mathbf{\hat{K}}$ are then used to calculate context-aware attention weights using the dot-product of both matrices. 
In contrast to the original implementation~\cite{yang2019context}, here we allow both $\lambda_{Q}^{h}$ and $\lambda_{K}^{h}$ to differ across heads, which allows variance in how context is integrated in each head.


We use a zero-symmetric gating unit to learn context gating factors $\{ \lambda_{Q}, \lambda_{K} \}$:
\begin{gather}
 \begin{bmatrix}
  \lambda_{Q}^{h} \\ 
  \lambda_{K}^{h}
   \end{bmatrix}
 = 
  \text{tanh} 
  \left(
  \begin{bmatrix}
  \mathbf{Q}^{h} \\
  \mathbf{K}^{h}
   \end{bmatrix} 
  \begin{bmatrix}
  \mathbf{V}_{Q}^{h} \\
  \mathbf{V}_{K}^{h}
   \end{bmatrix} + 
   \mathbf{C}^{h}
    \begin{bmatrix}
    \mathbf{U}_{Q} \\
    \mathbf{U}_{K}
   \end{bmatrix}
  \begin{bmatrix}
  \mathbf{V}_{Q}^{C} \\
  \mathbf{V}_{K}^{C}
   \end{bmatrix}
   \right)
\label{dot-qk}
\end{gather}
where $\{\mathbf{V}_{Q}^{h}, \mathbf{V}_{K}^{h}, \mathbf{V}_{Q}^{C}, \mathbf{V}_{K}^{C}\} \in \mathbb{R}^{d_{h} \times 1}$ are weights of linear layers to transform corresponding matrices. 
We chose to use \emph{tanh} as our activation function as this allows the context to contribute to $\mathbf{\hat{Q}}^{h}, \mathbf{\hat{K}}^{h}$ both positively and negatively\footnote{The original implementation~\cite{yang2019context} used \emph{sigmoid}.}. This enriches the representation space of both matrices, and the resulting attention distribution. We note that \emph{tanh} may increase the magnitude of $\mathbf{Q}$, $\mathbf{K}$, and large magnitude of $\mathbf{Q}$, $\mathbf{K}$ may push gradients to excessively small values, which may prevent model learning, as noted by~\citet{vaswani2017attention} and \citet{britz2017massive}. However, our results suggest that this did not negatively affect our model performance.

\subsection{Context-Guided \emph{Quasi}-Attention}
Our second neural network model (QACG-BERT) architecture proposes using a Quasi Attention function for (T)ABSA. The value of self-attention weights $\mathbf{A}_{\text{Self-Attn}}^{h}$ in a vanilla implementation (using \emph{softmax}), is bounded between $[0, 1]$. In other words, it only allows a convex weighted combination of hidden vectors at each position. This allows hidden states to contribute only additively, but not subtractively, to the attended vector. 
We include a \emph{quasi}-attention calculation to enable learning of both additive as well as subtractive attention~\cite{tay2019compositional}. Formally, we formulate our new attention matrix as a linear combination of a regular \emph{softmax}-attention matrix and a \emph{quasi}-attention matrix:
\begin{align}
    \mathbf{\hat{A}}^{h} = \mathbf{A}_{\text{Self-Attn}}^{h} +  
    \lambda_{A}^{h} \mathbf{A}_{\text{\emph{Quasi}-Attn}}^{h} 
    \label{quasi-attn}
\end{align}
where $\lambda_{A}^{h}$ is a scalar to represent the compositional factor to control the effect of context on attention calculation. $\mathbf{A}_{\text{Self-Attn}}^{h}$ is defined as in Eqn.~\ref{attn-self-attn}. To derive the quasi-attention matrix, we first define two terms quasi-context query $\mathbf{C}_{Q}^{h}$ and quasi-context key  $\mathbf{C}_{K}^{h}$:
\begin{gather}
 \begin{bmatrix}
  \mathbf{C}_{Q}^{h} \\ 
  \mathbf{C}_{K}^{h}
   \end{bmatrix}
 =    \mathbf{C}^{h}
    \begin{bmatrix}
    \mathbf{Z}_{Q} \\
    \mathbf{Z}_{K}
   \end{bmatrix}
\label{ctx-q-k}
\end{gather}
where $\{\mathbf{Z}_{Q}, \mathbf{Z}_{K}\} \in \mathbb{R}^{d_{e} \times d_{h}}$ are weights of linear layers to transform the raw context matrix, and $\mathbf{C}^{h}$ is the same context matrix in Eqn.~\ref{ctx-self-attn} (Defined in Sec.~\ref{sec:context}). Next, we define the quasi-attention matrix as:
\begin{align}
    \mathbf{A}_{\text{\emph{Quasi}-Attn}}^{h} = \alpha \cdot \text{sigmoid}\left(\frac{f_{\psi} (\mathbf{C}_{Q}^{h}, \mathbf{C}_{K}^{h})}{\sqrt{d_{h}}} \right) \label{attn-quasi-attn}
\end{align}
where $\alpha$ is a scaling factor and $f_{\psi}(\cdot)$ is a similarity measurement to capture similarities between $\mathbf{C}_{Q}^{h}$ and $\mathbf{C}_{K}^{h}$. For simplicity, we use dot-product to parameterize $f_{\psi}$, and set $\alpha$ to be $1.0$. Other $f_{\psi}$ that have been used include negative $L$-1 distance~\cite{tay2019compositional}. As a result, our $\mathbf{A}_{\text{Quasi-Attn}}$ is bounded between $[0, 1]$. We then derive our \emph{bidirectional} gating factor $\lambda_{A}$ as:
\begin{gather}
 \begin{bmatrix}
  \lambda_{Q}^{h} \\ 
  \lambda_{K}^{h}
   \end{bmatrix}
 = 
 \text{sigmoid} \left(
  \begin{bmatrix}
  \mathbf{Q}^{h} \\
  \mathbf{K}^{h}
   \end{bmatrix} 
  \begin{bmatrix}
  \mathbf{V}_{Q}^{h} \\
  \mathbf{V}_{K}^{h}
   \end{bmatrix} + 
    \begin{bmatrix}
    \mathbf{C}_{Q}^{h} \\
    \mathbf{C}_{K}^{h}
   \end{bmatrix}
  \begin{bmatrix}
  \mathbf{V}_{Q}^{C} \\
  \mathbf{V}_{K}^{C}
   \end{bmatrix}
   \right)
\label{dot-qk}
\end{gather}
\begin{align}
    \lambda_{A}^{h} = 1 - (\beta \cdot \lambda_{Q}^{h} + \gamma \cdot \lambda_{K}^{h})
\end{align} 
where $\{\beta, \gamma \}$ are scalars that control the composition weightings. For simplicity, we set $\{\beta, \gamma \}$ = 1.0. We formulate the gating factor to be \emph{bidirectional}, meaning it takes on both positive and negative values, and the output is bounded between $\{-1, 1\}$. Our intuition is that the context-based \emph{quasi}-attention may contribute either positively or negatively to the final attention weights. Consider Eqn.~\ref{quasi-attn}: as the first term $\mathbf{A}_{\text{Self-Attn}}^{h}$ is in $\{0, 1\}$, and the second term is made up of a term ($\lambda_{A}$) that is in $[-1,1]$ and another ($\mathbf{A}_{\text{Quasi-Attn}}$) that is in $[0, 1]$, hence the final attention $\mathbf{\hat{A}}$ lies in $[-1,2]$.
That is to say, the final attention weights can take values representing compositional operations including subtraction ($-$1), deletion ($\times$0), inclusion/addition ($+$1/$+$2) among hidden vectors across positions. 
We hypothesize that the \emph{quasi}-attention provides a richer method to integrate context into the calculation of attention.

\subsection{Classification}
We use the final hidden state (the output of the final layer of the BERT model) of the first classifier token (i.e., [CLS]) as the input to the final classification layer for a \textit{C}-class classification. This is similar to previous studies~\cite{sun2019utilizing}. For a given input sentence, we denote this vector as $e_{\text{CLS}} \in \mathbb{R}^{1 \times d}$. Then, the probability of each sentiment class $y$ is given by $y = \text{softmax}(e_{\text{CLS}}\mathbf{W}_{\text{CLS}}^{T})$ where $\mathbf{W}_{\text{CLS}} \in \mathbb{R}^{C \times d}$ are the weights of the classification layer, and $y \in \mathbb{R}^{1 \times C}$. The label with highest probability will be selected as the final prediction.

\begin{table*}[!t]
    \centering
    \scalebox{0.95}{
    \setlength\tabcolsep{5pt}
    \begin{tabular}{@{}lccccc@{}}
    \toprule
    \multirow{2}{*}{\textbf{Model}} & \multicolumn{3}{c}{Aspect Categorization} & \multicolumn{2}{c}{Sentiment}\\ \cmidrule(l){2-4} \cmidrule(l){5-6}
     & \multicolumn{1}{c}{Strict Accuracy /\%} &  \multicolumn{1}{c}{Macro-F1 /\%} & \multicolumn{1}{c}{AUC /\%} & \multicolumn{1}{c}{Accuracy /\%} & \multicolumn{1}{c}{AUC /\%} \\ \midrule
    LR~\cite{saeidi2016sentihood} & - &  39.3 & 92.4 & 87.5 & 90.5\\
    LSTM-Final~\cite{saeidi2016sentihood} & - & 68.9 &  89.8 & 82.0 & 85.4\\
    LSTM-Loc~\cite{saeidi2016sentihood} & - & 69.3 &  89.7 & 81.9 & 83.9\\
    SenticLSTM~\cite{ma2018targeted} & 67.4 & 78.2 & - & 89.3 & -\\
    Dmu-Entnet~\cite{liu2018recurrent} & 73.5 & 78.5 & 94.4 & 91.0 & 94.8\\
    BERT-single~\cite{sun2019utilizing} & 73.7 & 81.0 & 96.4 & 85.5 & 84.2\\
    BERT-pair~\cite{sun2019utilizing} & 79.8 & 87.9 & 97.5 & 93.6 & 97.0\\ \midrule
    CG-BERT (adapted from~\citealt{yang2019context}) & 79.7 (.3) & 87.1 (.2) & 97.5 (.2)  & 93.7 (.2) & 97.2 (.2) \\
    CG-BERT (auxiliary sentences) & 80.1 (.2) & 88.1 (.1) & 97.6 (.1)  & 93.1 (.3) & 97.0 (.2) \\
    QACG-BERT & 79.9 (.4) & 88.6 (.2) & 97.3 (.2) & \textbf{93.8} (.2) & \textbf{97.8} (.2) \\
    QACG-BERT (auxiliary sentences) & \textbf{80.9} (.3) & \textbf{89.7} (.2) & \textbf{97.8} (.2)  & 93.7 (.3) & 97.1 (.2) \\\bottomrule
    \end{tabular}
    }
    \vspace{3pt}
    \caption{Model performance on SentiHood TABSA dataset with best performances bolded. Top: Best results reported for previous models. ``-'' indicates not reported in the original paper. Bottom: Means and standard deviations across runs for our models. We also report our models trained with appended auxiliary sentences, the same inputs as~\citet{sun2019utilizing}.}
    \label{tab:sentihood}
\end{table*}

\subsection{Context Matrix} \label{sec:context}
We use a single integer to represent a context associated with an aspect and a target in any (T)ABSA task, and only an aspect in the ABSA task. We transform these integers into embeddings via a trainable embedding layer, which derives $\mathbf{C}^{h}$ in CG-BERT, and $\mathbf{C}_{Q}^{h}$ and $\mathbf{C}_{K}^{h}$ in QACG-BERT. For example, given $|t|$ targets and $|a|$ aspects for any (T)ABSA task, the total number of possible embeddings is $|t|\cdot|a|$. We then concatenate the context embedding with the hidden vector for each position $\mathbf{E} \in \mathbb{R}^{n \times d}$, and pass them into a feed-forward linear layer with a residual connection to formulate the context matrix $\mathbf{C}^{h} = [\mathbf{E}_{c}, \mathbf{E}]\mathbf{W}_{c}^{T}$ where $\mathbf{E}_{c} \in \mathbb{R}^{n \times d}$ is the context embedding and $\mathbf{W}_{c} \in \mathbb{R}^{d \times 2d}$ are the learnt weights for this feed-forward layer.

\subsection{Integration with Pretrained BERT} \label{sec:init}
Previous studies show that fine-tuning pretrained BERT models increases performance significantly in many NLP tasks~\cite{sun2019utilizing, rietzler2020adapt}. Since our models share most of the layers with a standard BERT model, we import weights from pretrained BERT models for these overlapping layers. The weights of the newly added layers are initialized to be small\footnote{We also tried initializing the weights in the newly added layers with larger variance, and found similar performance.} random numbers drawn from a normal distribution $\mathcal{N}(0, \sigma^{2})$ with $\sigma = e^{-3}$. As a result, the gating factors in Eqn.~\ref{ctx-self-attn} and Eqn.~\ref{attn-self-attn} start at values close to zero. This initialization enables the task-specific weights to start from the pretrained weights and slowly diverge during training.

%% file: 4_experiment.tex
\section{Experiments}

\begin{table*}[!t]
    \centering
    \scalebox{0.95}{
    \setlength\tabcolsep{5pt}
    \begin{tabular}{@{}lcccccc@{}}
    \toprule
    \multirow{2}{*}{\textbf{Model}} & \multicolumn{3}{c}{Aspect Categorization} & \multicolumn{3}{c}{Sentiment}\\ \cmidrule(l){2-4} \cmidrule(l){5-7}
     & \multicolumn{1}{c}{Precision /\%} &  \multicolumn{1}{c}{Recall /\%} & \multicolumn{1}{c}{F1 /\%} & \multicolumn{1}{c}{Binary /\%} & \multicolumn{1}{c}{3-class /\%} & \multicolumn{1}{c}{4-class /\%} \\ \midrule
    XRCE~\cite{brun2014xrce} & 83.23 & 81.37 & 82.29 & - & - & 78.1 \\
    NRC-Canada~\cite{kiritchenko2014nrc} & 91.04 & 86.24 & 88.58 & - & - & 82.9 \\
    BERT-single~\cite{sun2019utilizing} & 92.78 & 89.07 & 90.89 & 93.3 & 86.9 & 83.7 \\
    BERT-pair~\cite{sun2019utilizing} & 93.57 & 90.83 & 92.18 & 95.6 & 89.9 & 85.9 \\ \midrule
    CG-BERT (adapted from~\citealt{yang2019context}) & 93.02 (.27) & 90.00 (.33) & 91.49 (-) & 94.3 (.3) & 89.9 (.2) & 85.6 (.4) \\
    CG-BERT (auxiliary sentences) & 93.12 (.37) & 90.17 (.23) & 91.62 (-) & 94.7 (.3) & 90.1 (.4) & 85.7 (.5) \\
    QACG-BERT & \textbf{94.38 (.31)} & \textbf{90.97 (.28)} & \textbf{92.64 (-)} & 95.6 (.4) & 90.1 (.3) & 86.8 (.8) \\
    QACG-BERT (auxiliary sentences) & 94.27 (.39) & 90.12 (.34) & 92.14 (-) & \textbf{95.8 (.3)} & \textbf{90.4 (.5)} & \textbf{86.9 (.7)} \\ \bottomrule
    \end{tabular}
    }
    \vspace{3pt}
    \caption{Model performance on the Semeval-2014 Task 4 ABSA dataset, with best performances bolded. Aspect categorization and sentiment classification corresponds to Subtask 3 and Subtask 4 respectively. Top: Best results reported for previous models. ``-'' indicates not reported in the original paper. Bottom: Means and standard deviations across runs for our models.}
    \label{tab:semeval}
\end{table*}

\subsection{Datasets}
We evaluate our models with two datasets in English. For the TABSA task, we used the Sentihood dataset~\footnote{\url{https://github.com/uclnlp/jack/tree/master/data/sentihood}} which was built by questions and answers from Yahoo! with location names of London, UK. It consists of 5,215 sentences, with 3,862 sentences containing
a single target and 1,353 sentences containing multiple targets. For each sentence, we predict sentiment label $y$ for each target-aspect pair $(t, a)$. For the ABSA task, we used the dataset from SemEval 2014, Task 4~\footnote{\url{http://alt.qcri.org/semeval2014/task4/}}, which contains 3,044 sentences from restaurant reviews. For each sentence, we predict the sentiment label $y$ for each aspect $a$. 
Each dataset is partitioned to train, development and test sets as in its original paper.

As in previous studies~\cite{pontiki2014semeval, saeidi2016sentihood}
we define two subtasks for each dataset: (1) aspect categorization and 2) aspect-based sentiment classification. For aspect categorization, the problem is to detect whether a aspect $a$ is mentioned (i.e., \emph{none} means not mentioned) in the input sentence for a target $t$ if it is a TABSA task. For aspect-based sentiment classification, we give the model aspects that present (i.e., ignoring \emph{none}'s) and have the model predict the valence of the sentiment (i.e., potential labels include \emph{negative} and \emph{positive} for Sentihood, and \emph{negative}, \emph{neutral}, \emph{positive}, \emph{conflicting sentiment} for Semeval Task 4).


\subsection{Experiment Settings}
As in the original BERT-base model~\cite{devlin2019bert}, our models consists of 12 heads and 12 layers, with hidden layer size 768. The total number of parameters for both of our models increased slightly due to the additional linear layers added comparing to previous BERT-based models for ABSA tasks~\cite{sun2019utilizing} which consists of about 110M parameters. The CG-BERT and QACG-BERT consists of about 124M parameters. We trained for 25 epochs with a dropout probability of 0.1. The initial learning rate is $2e^{-5}$ for all layers, with a batch size of 24. We used the pretrained weights from the uncased BERT-base model~\footnote{\url{https://storage.googleapis.com/bert_models/2020_02_20/uncased_L-12_H-768_A-12.zip}}.

We used a single Standard NC6 instance on Microsoft Azure, which is equipped with a single NVIDIA Tesla K80 GPU with 12G Memory. Training both models across two datasets took approximately 11 hours.

\begin{figure}[!htb]
    \centering
    \includegraphics[width=0.48\textwidth]{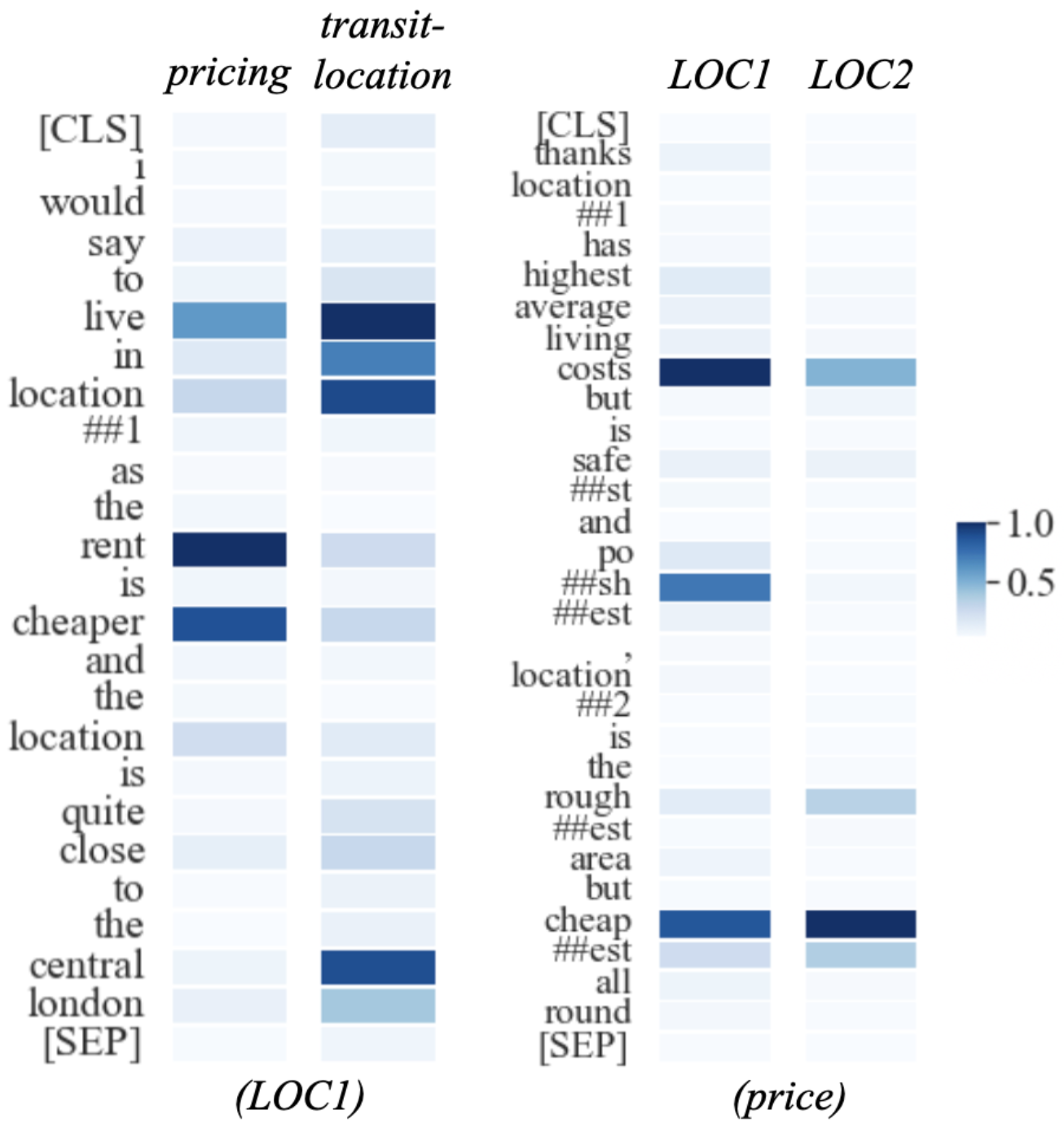}
    \caption{Examples of relevance scores via gradient sensitivity analysis for different aspects and locations, from our QACG-BERT model on the SentiHood TABSA test set. Left: Gradients associated with two aspects \{\emph{pricing}, \emph{transit-location}\} for target \emph{LOC1}. Right: Gradients associated with the \emph{general} aspect, for two targets \emph{LOC1} and \emph{LOC2}. Values are normalized with respect to the maximum value within a sentence.}
    \label{fig:model}
\end{figure}

\subsection{Exp-I: TABSA}
For the TABSA task, we compared the performance of our models with previous models in Table~\ref{tab:sentihood}.
Following~\citet{ma2018targeted} and~\citet{sun2019utilizing}, for aspect categorization (is a given aspect present in the sentence? If aspect is not present, the label is by definition \emph{none}), we report strict accuracy (model needs to correctly identify all aspects for a given target in the sentence to be counted as accurate), Macro-F1 (the harmonic mean of the Macro-precision and Macro-recall of all targets.) and AUC. For sentiment classification (given an aspect present in the sentence, is the valence \emph{negative} or \emph{positive}?), we report accuracy and AUC. 


\paragraph{Results} 
Our results showed that modifying BERT to be context-aware resulted in increased performance, surpassing the previous state-of-the-art. 
Across multiple evaluation metrics, our proposed quasi-attention model with pretrained BERT weights (QACG-BERT) performed the best. Additionally, we also evaluated our models trained with inputs appended with auxiliary sentences as in~\cite{sun2019utilizing} for comparison. Specifically, we append the target and aspect, for example, ``[SEP] - location - 1 - price'', to the input. Our results showed that for some metrics, using auxiliary sentences improves performance but not in others.



\begin{figure*}[!htb]
\minipage{0.98\textwidth}
  \includegraphics[width=\linewidth]{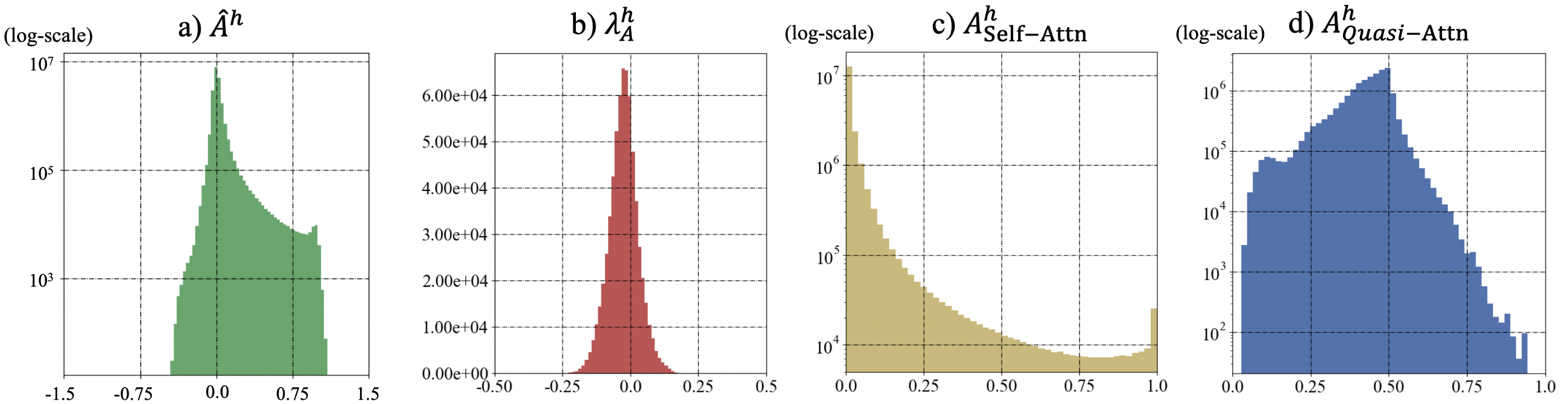}
\endminipage
\caption{Histogram of QACG-BERT weights on $n=$200 randomly selected examples from SentiHood test set where true label is either \emph{negative} or \emph{positive}. For the matrices, each value in the matrix is a data point in the histogram. We note that the vertical axes of (a), (c) and (d) are in log-scale.}
\label{fig:attn-dist}
\end{figure*}

\begin{figure}[!htb]
    \centering
    \includegraphics[width=0.38\textwidth]{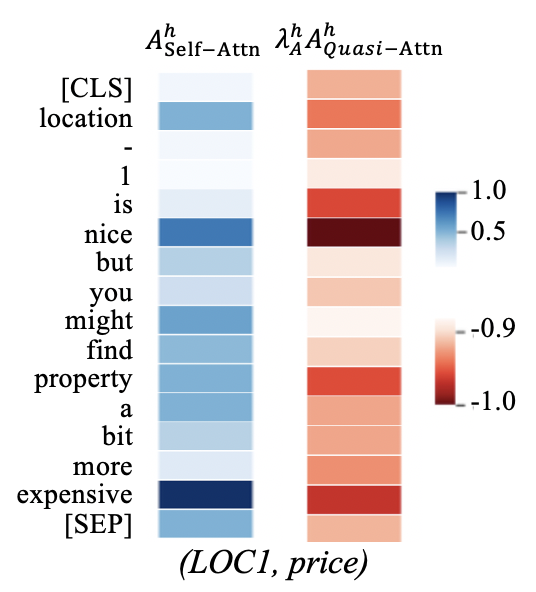}
    \caption{Example of attention weights extracted from different matrices for an aspect-location pair, (\emph{LOC1}, \emph{price}), from the SentiHood test set. Values are normalized with respect to the absolute maximum value within a sentence. We note the high negative quasi-attention on \textbf{nice}, which is often a positive word, but not in the context of \emph{price}.}
    \label{fig:qa-attn-viz}
\end{figure}

\subsection{Exp-II: ABSA}

For the ABSA task, we compared our models with multiple best performing models for the SemEval-2014 Task 4 dataset in Table~\ref{tab:semeval}. 
Following~\citet{pontiki2016semeval}, for aspect categorization, we report Precision, Recall, and F1. For aspect-based sentiment classification, we report accuracies for three different evaluations: binary classification (i.e., \emph{negative} or \emph{positive}), 3-class classification (\emph{negative}, \emph{neutral} or \emph{positive}) and 4-class classification ( \emph{negative}, \emph{neutral}, \emph{positive} or \emph{conflict}).

\paragraph{Results} Consistent with Exp-I, our models improved over previous state-of-the-art models on the SemEval-2014 Task 4 dataset. For aspect catogorization, our CG-BERT performed slightly worse than previous best performing models while QACG-BERT performed surpass the best model in F1 scores. 
For aspect sentiment classification, QACG-BERT performs better than CG-BERT and both models surpass previous SOTA performance. Similar to TABSA, our results show slight drops in some metrics while maintain same performance in others when trained with appended auxiliary sentences. 


\subsection{Feature Importance Analysis} \label{sec:div-grad}

To visualization importance of input tokens, we conduct gradient sensitivity analysis~\cite{li2016visualizing, arras2017explaining}. In Fig.~\ref{fig:model}, we visualize the gradient scores of our QACG-BERT model over two input sentences for which our model predicts sentiment labels correctly. For the first example, words like \textbf{rent} and \textbf{cheaper} are most relevant to the \emph{price} aspect while words like \textbf{live}, \textbf{location}, \textbf{central} are more relevant to the \emph{transit-location} aspect. Thus, our model learns how to pick out words that are important under different contexts. 
The second example concerns two targets (i.e., locations) for the same aspect, \emph{price}, where the sentiment label for \emph{LOC1} is \emph{negative} while for \emph{LOC2} is \emph{positive}. 
For \emph{LOC1}, the model correctly identifies \textbf{costs} (which in context refers to \emph{LOC1}) and \textbf{posh}\footnote{The BERT Tokenizer breaks up `poshest' into po-sh-est. We note that the BERT vocabulary does not have `posh' in it, but after fine-tuning, appears to learn that the word is price-relevant.}. By contrast, the model identifies \textbf{cheap} when given \emph{LOC2} as context, and assigns a greater gradient value, compared to the same word for \emph{LOC1}. As a result, the model is able to identify words corresponding to different targets and different aspects.

\subsection{Quasi-attention Visualization}

We inspected the quasi-attention parameters learnt by our models. Specifically, we took the QACG-BERT model trained on the Sentihood dataset and extracted values for the following four variables: the final attention weights matrix $\mathbf{\hat{A}}^{h}$, the \emph{bidirectional} gating factor matrix $\lambda_{A}^{h}$, the vanilla self-attention matrix $\mathbf{A}_{\text{Self-Attn}}^{h}$ and the context-guided quasi-attention matrix $\mathbf{A}_{\text{\emph{Quasi}-Attn}}^{h}$. Fig.~\ref{fig:attn-dist} illustrates the histogram of values drawn from 200 examples from the test set.

We made several key observations. First, the behaviour of $\lambda_{A}^{h}$ follows our intuition; it acts as a bidirectional control gate, with slightly more negative values, and determines whether context contributes to attention positively or negatively. Second, the learnt weights in $\mathbf{A}_{\text{\emph{Quasi}-Attn}}^{h}$ are not near zero, with the mass of the distribution between .25 and .50, thus, it does contribute to attention. 
Lastly, the non-zero weights in the final matrix $\mathbf{\hat{A}}^{h}$ are mainly positive, but some of the weights take on negative values due to the bidirectional gating factor. This is important as it enables the model to attend to and ``de-attend from" different parts of the input.

Finally, we turn to visualizing the quasi-attention. In Fig.~\ref{fig:qa-attn-viz}, we visualize the weights of $\mathbf{A}_{\text{Self-Attn}}^{h}$ and the product $\lambda_{A}^{h} \mathbf{A}_{\text{\emph{Quasi}-Attn}}^{h}$ extracted from one of the heads in the first self-attention layer, over an example input sentence from the test set. For this sentence, our model correctly predicted the sentiment label $\emph{negative}$ for the \emph{price} aspect of \emph{LOC1}. 
In this example, \textbf{nice} is a positive word for aspects like \emph{general}. However, with respect to \emph{price}, \textbf{nice} actually suggests a higher cost and thus is more negative. We see that while $\mathbf{A}_{\text{Self-Attn}}^{h}$ on \textbf{nice} is high, $\mathbf{A}_{\text{Quasi-Attn}}^{h}$ on \textbf{nice} is negative, which ``subtracts'' from the final attention. 
As a result, the sum derived by Eqn.~\ref{quasi-attn} makes the final attention weight on the word \textbf{nice} less positive. We note that empirically, we find that the total attention $\mathbf{\hat{A}}^{h}$ is usually positive (Fig.~\ref{fig:attn-dist}(a)), i.e., quasi-attention, when negative, tends to be smaller in magnitude than self-attention.

%% file: 5_conclusion.tex
\section{Conclusion}
We proposed two context-aware BERT-based models for ABSA and TABSA, which outperformed state-of-the-art results on two datasets. Our first CG-BERT model introduced a way of integrating context into pretrained BERT for ABSA. The second QACG-BERT model allowed quasi-attention, which enables compositional attention including subtraction ($-$1), deletion ($\times$0), inclusion/addition ($+$1/$+$2). Our results and analyses show strong performance results, especially for our QACG-BERT model, in solving (T)ABSA tasks, and suggest potential success for such context-aware mechanisms for other context-based tasks in NLP.

%% file: 6_acknowledgement.tex
\section{Acknowledgements}
This work was supported in part by a Singapore Ministry of Education Academic Research Fund Tier 1 grant to DCO.
%